\documentclass[10pt,twocolumn,letterpaper]{article}

\usepackage{cvpr}              % To produce the CAMERA-READY version
\usepackage{graphicx}
\usepackage{amsmath}
\usepackage{amssymb}
\usepackage{booktabs}
\newcommand\blfootnote[1]{%
  \begingroup
  \renewcommand\thefootnote{}\footnote{#1}%
  \addtocounter{footnote}{-1}%
  \endgroup
}

\usepackage[pagebackref,breaklinks,colorlinks]{hyperref}
\usepackage[misc]{ifsym}

\usepackage[capitalize]{cleveref}
\crefname{section}{Sec.}{Secs.}
\Crefname{section}{Section}{Sections}
\Crefname{table}{Table}{Tables}
\crefname{table}{Tab.}{Tabs.}

\begin{document}
\blfootnote{$^\text{\Letter}$ Corresponding author} 
%%%%%%%%% TITLE - PLEASE UPDATE
\title{IC-Portrait: In-Context Matching for View-Consistent Personalized Portrait Generation}

\author{Han Yang\textsuperscript{1,2}\,\,\,\,
Enis Simsar\textsuperscript{1}\,\,\,\,
Sotiris Anagnostidis\textsuperscript{1}\,\,\,\,  
Yanlong Zang\textsuperscript{3}\,\,\,\, 
Thomas Hofmann\textsuperscript{1}\,\,\,\,
Ziwei Liu\textsuperscript{4}$^\text{\Letter}$\,\,\,\,\\
\centerline{
\textsuperscript{1}ETH Zurich,\,\,\,\,\
\textsuperscript{2}ZMO AI Inc.}\\
\centerline{\textsuperscript{3}Zhejiang University\,\,\,\,\,
\textsuperscript{4}S-Lab, Nanyang Technological University}\\
{\tt\small hanyang@ethz.ch, ziwei.liu@ntu.edu.sg}
% For a paper whose authors are all at the same institution,
% omit the following lines up until the closing ``}''.
% Additional authors and addresses can be added with ``\and'',
% just like the second author.
% To save space, use either the email address or home page, not both
}

\begin{figure}[htb]

\twocolumn[{
\renewcommand\twocolumn[1][]{#1}%
\maketitle
\vspace{-32pt}

\begin{center}
  \centering
  \includegraphics[width=0.95\textwidth]{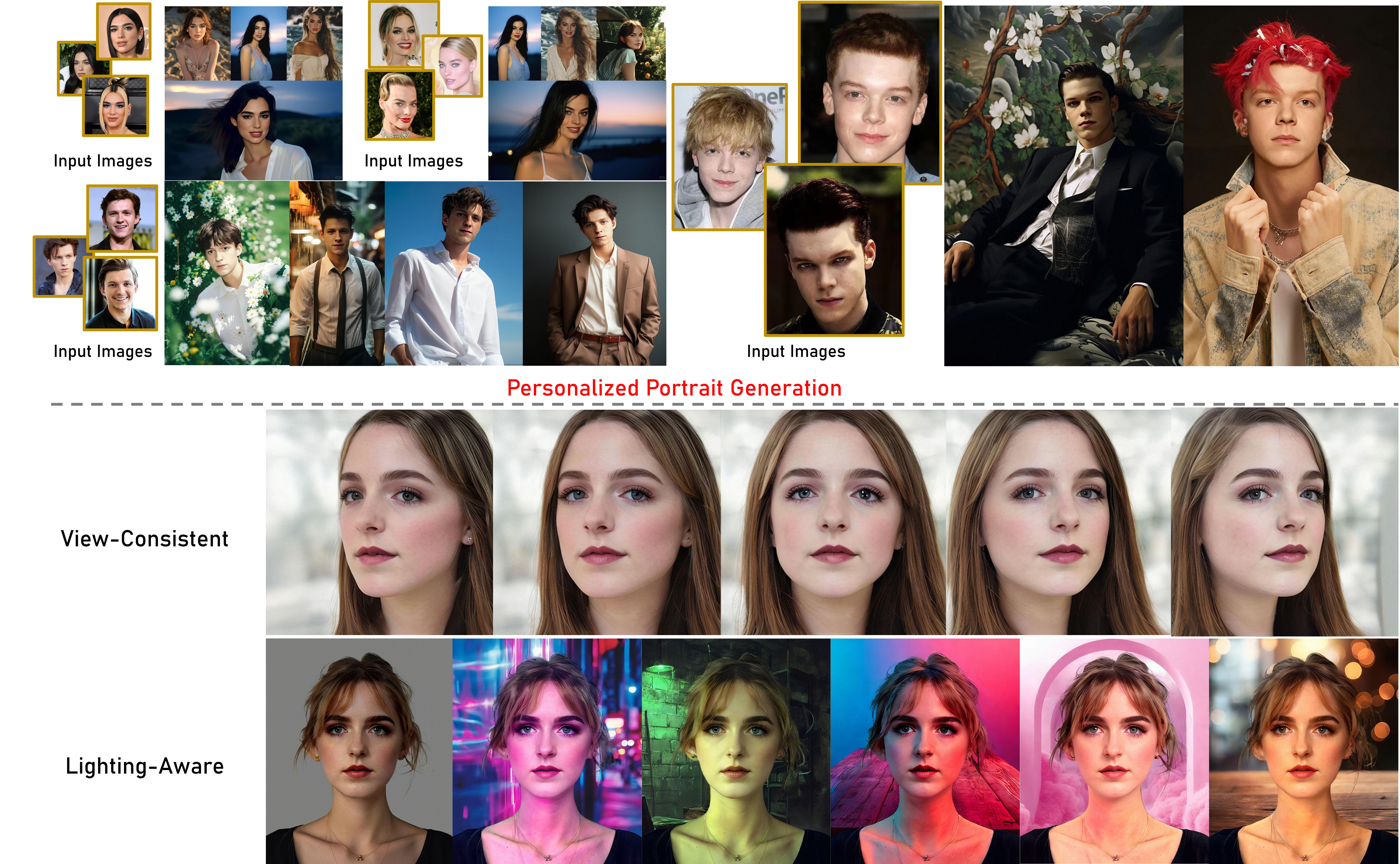}
  \vspace{-20pt}

\end{center}

\caption{IC-Portrait introduces an innovative method for creating photorealistic portrait images. By incorporating style reference images, our system generates personalized portraits that maintain consistent viewpoints and accurately respond to lighting conditions.}
\label{fig:teaser}
}]
\end{figure}

%%%%%%%%% ABSTRACT
\begin{abstract}

Existing diffusion models show great potential for identity-preserving generation. However, personalized portrait generation remains challenging due to the diversity in user profiles, including variations in appearance and lighting conditions. To address these challenges, we propose \textbf{IC-Portrait}, a novel framework designed to accurately encode individual identities for personalized portrait generation. Our key insight is that pre-trained diffusion models are fast learners (\textit{e.g.,} $100 \sim 200$ steps) for in-context dense correspondence matching, which motivates the two major designs of our IC-Portrait framework. Specifically, we reformulate portrait generation into two sub-tasks: \textbf{1) Lighting-Aware Stitching}: we find that masking a high proportion of the input image, \textit{e.g.,} 80\%, yields a highly effective self-supervisory representation learning of reference image lighting. \textbf{2) View-Consistent Adaptation}: we leverage a synthetic view-consistent profile dataset to learn the in-context correspondence. The reference profile can then be warped into arbitrary poses for strong spatial-aligned view conditioning. Coupling these two designs by simply concatenating latents to form ControlNet-like supervision and modeling, enables us to significantly enhance the identity preservation fidelity and stability.
Extensive evaluations demonstrate that IC-Portrait consistently outperforms existing state-of-the-art methods both quantitatively and qualitatively, with particularly notable improvements in visual qualities. Furthermore, IC-Portrait even demonstrates 3D-aware relighting capabilities.

%Current portrait generation methods based on ID embedding fail to capture local details with non-spatial vector representation, and those relying on fine-tuning struggle to learn the precise identity from diverse profile images. To alleviate these problems, we adopt the idea of novel view synthesis; instead of regenerating a somewhat similar portrait, it readjusts a user's profile photo to match a given pose of a style reference image which gives the template portrait for consistent identity representation. For relightable stitching, it devises techniques to blend a user's profile and novel view into a style reference image despite differences in lighting and color. More importantly, we inherently integrate these sub-processes into one end-to-end framework rather than assembling them in an ad-hoc manner.

%To give the capability of novel view synthesis to ControlNet, NovelPortrait refactors the ControlNet structure through dual condition learning in the form of spatial self-attention. This enables it to learn novel-view transformation and relightable adaptation simultaneously while still applicable to customized model such as Dreambooth. We coherently combine the advantages of embedding-based models, Dreambooth fine-tuning techniques, and self-attention references, resulting in flexibility during actual deployment.
\href{https://www.youtube.com/watch?v=By-y3IH6ejM}{ Please check our video}.

\end{abstract}

%%%%%%%%% BODY TEXT

% Some commands used in this file

\section{Introduction}

Portrait generation has emerged as a pivotal area of research in computer graphics, driven by its applications in digital content creation, virtual avatars, gaming, and augmented reality. 
Although recent advances in portrait generation models have enabled increasingly realistic and visually compelling portraits, fundamental challenges persist in creating images that maintain precise identity while adapting to diverse artistic styles and environmental conditions. 

Portrait generation approaches generally fall into two categories: 1) \textbf{prompt-based stylization} and 2) \textbf{reference image-based} portrait generation. Prompt-based stylization~\cite{IP-Adapter,photomaker,InstantID} attempts to preserve the subject's identity while allowing users to guide the generation via descriptive text prompts. The main benefit of this setting is the ability to edit the style according to the user's preference. In contrast, reference image-based portrait generation~\cite{easyphoto,Megaportrait} directly leverages existing images to define the desired style, changing only the identity, and providing a more intuitive interface for users. We base our work on the latter approaches due to the higher-quality images they result to.

\begin{figure}[thb]
  \centering
   \includegraphics[width=0.9\linewidth]{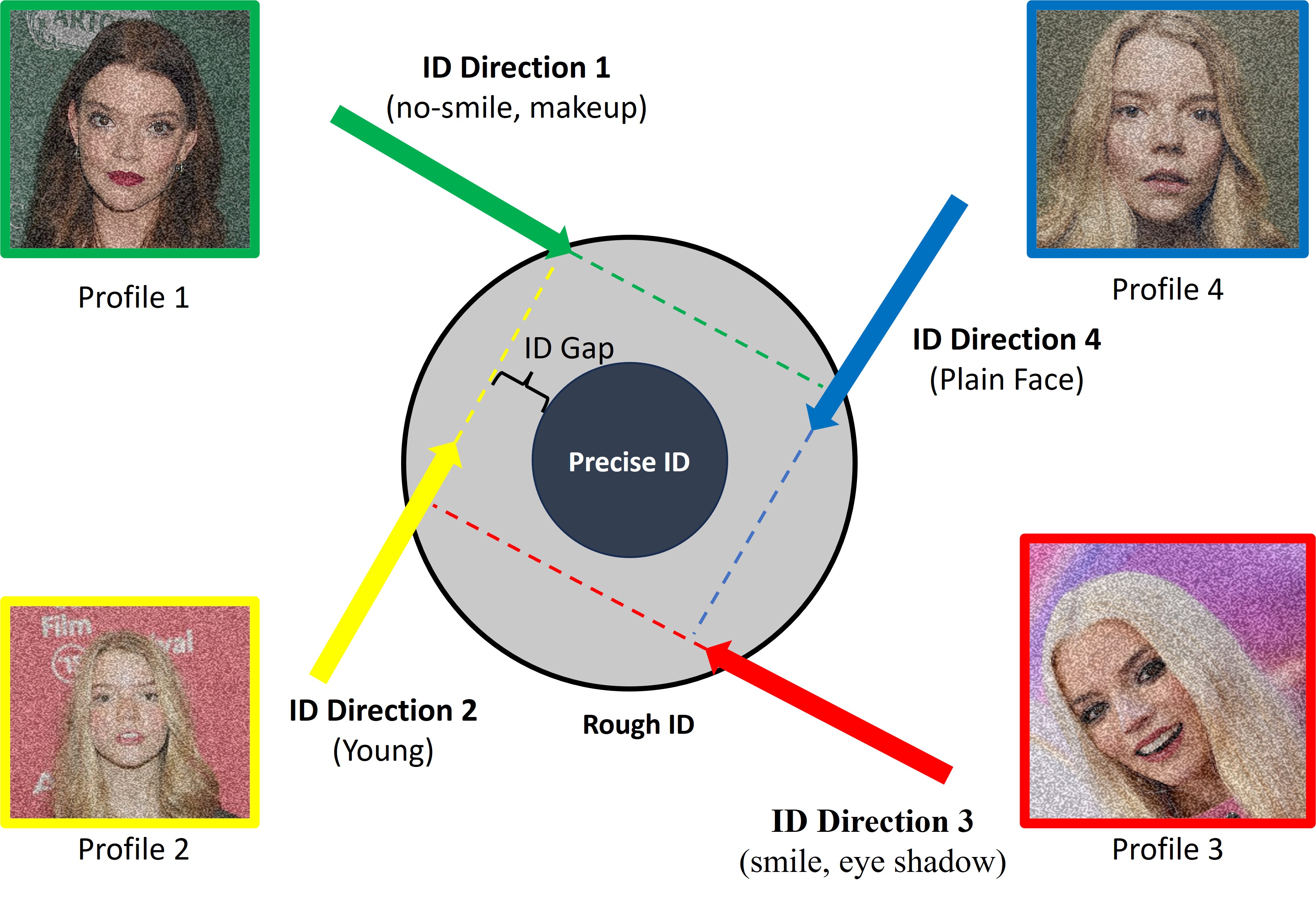}
   \caption{Different expressions and makeup are presented at different profile images showcasing the diversity of human profiles.}
   \label{fig:diverse}
\end{figure}

%Problem Statement
Despite progress in portrait generation, AI-generated portraits still fail to completely replace offline photography. Given the aforementioned problems, we ask: 
\begin{center}
    {\it  What factors prevent portrait-generation methods from achieving nearly exact identity resemblance?} 
\end{center}   
One primary issue is managing diverse user photos, which we term \textbf{intra-identity diversity}. User-provided profile images can vary significantly in terms of shooting conditions, such as lighting (ranging from bright sunlight to dim indoor light), facial expressions (from extreme joy to deep sadness), and makeup (from natural-looking to heavily made-up). These variations introduce a high degree of uncertainty in identity (ID) learning, as also illustrated in Fig.~\ref{fig:diverse}, and directly impact the quality and accuracy of generated portraits. Traditional optimization methods like DreamBooth~\cite{Dreambooth} may converge toward a possible identity representation without capturing the precise identity characteristics. Such solutions fail to convince the human brain, which processes facial information with remarkable efficiency through dedicated neural pathways \cite{faceperception}, and thus can sensitively distinguish the degraded ID quality in 
personalized portrait generation, which demands high ID-consistency for portrait generation methods.
% it becomes extremely difficult for models to accurately extract and represent the unique identity of an individual.

\label{sec:challenges}

Another major issue are the potential differences between user-provided profile images and style reference images, which we call \textbf{adaptation gap}. Different portrait styles come with distinct lighting, shading, and color characteristics. Attempting to merge the identity from the profile image with the style of the reference image presents a fundamental challenge to ensure both a high level of resemblance to the person in the profile and a seamless visual integration with the style reference.

% \textbf{Face Perception Sensitivity}.Human brain has a dedicated zone to parse face information~\cite{faceperception}: a glance at a person's face, lasting merely a split-second, reveals a wealth of information. While humans can easily recognize a person despite varying lighting, expressions, and cosmetics, AI models face a significant challenge. They must analyze pixel-level details to distinguish identity-inherent features from those due to the shooting context and then generate portraits that preserve the precise identity and adapt to different artistic styles.

% \textbf{Managing Diverse User Photos}.  User photos vary widely due to different shooting conditions and facial attributes like extreme expressions and significant makeup differences. This leads to high uncertainty in identity (ID) learning, as shown in Fig.~\ref{fig:diverse}, and directly impacts the quality and accuracy of generated portraits. Especially for gradient descent-based optimization method such as Dreambooth~\cite{Dreambooth}, the gradient points at a direction towards one possible identity representation but not to the precise ID. 

% \textbf{Gap between Profiles and Style References}: Diverse portrait styles have diverse lighting and shading. It is extremely challenging to meet the standard of resemblance and coherence simultaneously due to the significant gap between profile images and style reference images. Profile images may have their own specific lighting conditions, facial expressions, and other characteristics that are different from those of style reference images.

Given the analysis above, our solution is IC-Portrait, a novel framework designed to generate images with precise ID characteristics while ensuring lighting consistency. IC-Portrait follows the reference image-based portrait generation methods and achieves lossless face/head swapping. IC-Portrait reformulates portrait generation into two complementary sub-tasks: \textbf{lighting-aware stitching} and \textbf{view-consistent adaptation}. We elaborate on how the reformulation helps solve the \textit{intra-identity diversity} and \textit{adaptation gap} issues as follows:
\begin{enumerate}
    \item For lighting-aware stitching, we employ a high-proportion masked autoencoding technique of the input image, motivated and inspired by Masked Autoencoders (MAE)~\cite{MAE}. This technique enables self-supervised learning of lighting characteristics from style reference images. By doing so, the \textbf{adaptation gap} between profile and reference images is effectively reduced with well-preserved local lighting clues and global shading effect.
    \item For view-consistent adaptation, we leverage a synthetic view-consistent profile dataset~\cite{Single-Image,SDXL,EG3D}, that allows the model to learn in-context correspondence, enabling it to warp the reference profile into different poses. This provides robust spatial-aligned view conditioning, crucial for maintaining consistency across varying viewpoints. We can thus achieve copy-paste level consistency of a specific chosen profile to fully reconstruct the target identity in our portrait result. % In plain words: 
    \begin{center}
        {\it For each generation, the portrait only needs to resemble one specific profile picture, not all of them simultaneously.}
    \end{center}
    In this manner, instead of learning the universal ID representation of a person, we simply attempt to re-render the appearance in the style of the reference image, thereby creating a relaxed yet smart setting with the aim of bypassing the \textbf{intra-identity diversity} problem.
\end{enumerate}

Our main contributions can be summarized as follows: 1) We introduce IC-Portrait, a diffusion-based framework for view-consistent personalized portrait generation. By refactoring the portrait generation into two sub-processes, view-consistent adaptation, and lighting-aware stitching, it can not only produce almost visually lossless ID preservation but also enable lighting-invariant and view-consistent portrait generation. 2) To enable such decomposition, we find: i) pre-trained diffusion models are fast learners for in-context correspondence, which is utilized as spatial-aligned control signals for view-consistent generation. ii) high-proportion masked autoencoding (around 80\%) serves as an effective identity-agnostic representation to leak lighting information in a non-trivial self-supervisory task.
3) To utilize the correspondence and lighting representation, we curate a synthetic dataset with multi-view portrait images of 35200 view pairs triggered by collected celebrity names and implement the appearance net as well as the matching net with conceptually simple ControlNet-like structure; the appearance reference and the masked image are concatenated horizontally forming in-context scenario.
4) Through extensive evaluations, we demonstrate that IC-Portrait outperforms current state-of-the-art methods both quantitatively and qualitatively. IC-Portrait achieves 0.674 for Arcface~\cite{arcface} ID metric and significantly better visual consistency in portrait generation, highlighting its effectiveness.

%1) An insightful perspective is proposed by modeling the portrait generation as a relightable novel-view synthesis problem. 2) We design a dual condition learning framework with a ControlNet-like model which preserves the fine plug-and-play ability of controlnet while enabling copy-level texture consistency. 3) We introduce a cross-profile training scheme in the personalization training phase, which enhances generalization ability. We can thus flexibly adjust the stacked embedding to achieve different levels of trade-off in terms of ID similarity and style reference. 4) IC-Portrait achieves unprecedented ID similarity compared to existing methods. For example, when compared to Method PhotoMaker~\cite{photomaker}, IC-Portrait shows a 20\% higher ID similarity in quantitative experiments, demonstrating its superiority in maintaining the identity of the profile image.

\section{Related Works}

% \subsection{Latent Diffusion}
% Latent Diffusion Models (LDMs) \cite{LatentDiffusion} as well as its derivative Stable Diffusion \cite{LatentDiffusion} 
\noindent \textbf{Latent Diffusion.} Latent Diffusion Models (LDMs) \cite{LatentDiffusion} and their derivative, Stable Diffusion \cite{LatentDiffusion}, have been game changers in generative modeling methods. Before LDMs, diffusion models were considered slow and computationally expensive as they operated on the pixel space in a Markovian fashion. LDM first encodes the image into latent space with a variational autoencoder (VAE) \cite{VAE}, which significantly reduces the resolution and trains the diffusion model in the latent space. LDM also proposes a cross-attention paradigm by introducing the cross-attention mechanism into the Unet \cite{LatentDiffusion} structure, enabling text-controllable generation. We also adopt latent diffusion as our backbone.
\noindent \textbf{Human-centric Image Generation.}
Traditional techniques such as face-swap \cite{faceshifter,simswap,Hififace} mainly focused on replacing facial features without truly capturing the unique identity of the source. The emergence of diffusion models has radically transformed this approach, enabling precise generation of the target individual based on a given style or prompt. Encoder-driven approaches, such as \cite{IP-Adapter,FastComposer,photomaker}, offer improvements by utilizing image encoders to extract identity features. However, these methods face challenges when it comes to adapting to specific style references. Optimization-based solutions such as Dreambooth~\cite{Dreambooth} and LoRA~\cite{Lora} can better maintain the identity of the subject but often fall short in delivering the variety of aesthetic styles due to overfitting and thus require intricate regularization design \cite{Dreambooth}. Our approach can be compatible with off-the-shelf personalized models, thus avoiding further fine-tuning. Notably, we will also present a better practice of training customized diffusion models following the DreamBooth~\cite{Dreambooth} setting.

%On the consumer side, Remini \cite{remini} stands out as one of the top-selling AI portrait enhancement apps on platforms like the App Store and Google Play. While it's not a core part of our technical analysis, it's worth noting its market presence as an example of a consumer-facing application in this space.

\noindent \textbf{Novel-view Synthesis}
Earlier methods mainly focused on using 3D Morphable Models (3DMM) \cite{3DMM} to explicitly reconstruct human faces. However, the limited parametric space of 3DMM hinders the expressiveness of diverse facial details. Neural Radiance Fields (NeRF) \cite{NeRF} demonstrate promising results by learning the implicit representation of the target object. %FDNeRF \cite{FDNeRF} can leverage the cross-expression features to reconstruct the facial NeRF representation with only a few profile images. 

\section{Methodology}

\subsection{Motivation}

Portrait generation has faced significant challenges in creating personalized, high-quality images. Traditional ID-embedding approaches often sacrifice spatial detail, while DreamBooth-style test-time optimization methods struggle to generate diverse representations of the same identity. Drawing inspiration from Novel View Synthesis (NVS), which excels at identity preservation, we recognized that effective portrait generation should prioritize the accurate reproduction of individual reference images rather than pursuing a generalized model. However, while NVS effectively handles facial reposing, it lacks the flexibility to adapt to different artistic styles and lighting conditions. To address this limitation, we developed IC-Portrait, which incorporates a dedicated relighting phase. Our framework moves beyond simple copy-paste-harmonize strategies, instead of implementing an end-to-end system that maintains consistency across viewpoint changes and lighting conditions.

% \begin{figure}[thb]
%   \centering
%    \includegraphics[width=1\linewidth]{images/multi-view.jpg}
%    \caption{The synthetic multi-view dataset is used to train our appearance and matching nets. We employ SDXL to generate human images by randomly selecting celebrity names. Then, we reproject the image into the latent space and generate multi-view images using EG3D.}
%    \label{fig:multi-view}
% \end{figure}

\begin{figure*}[thb]
  \centering
   \includegraphics[width=0.9\linewidth]{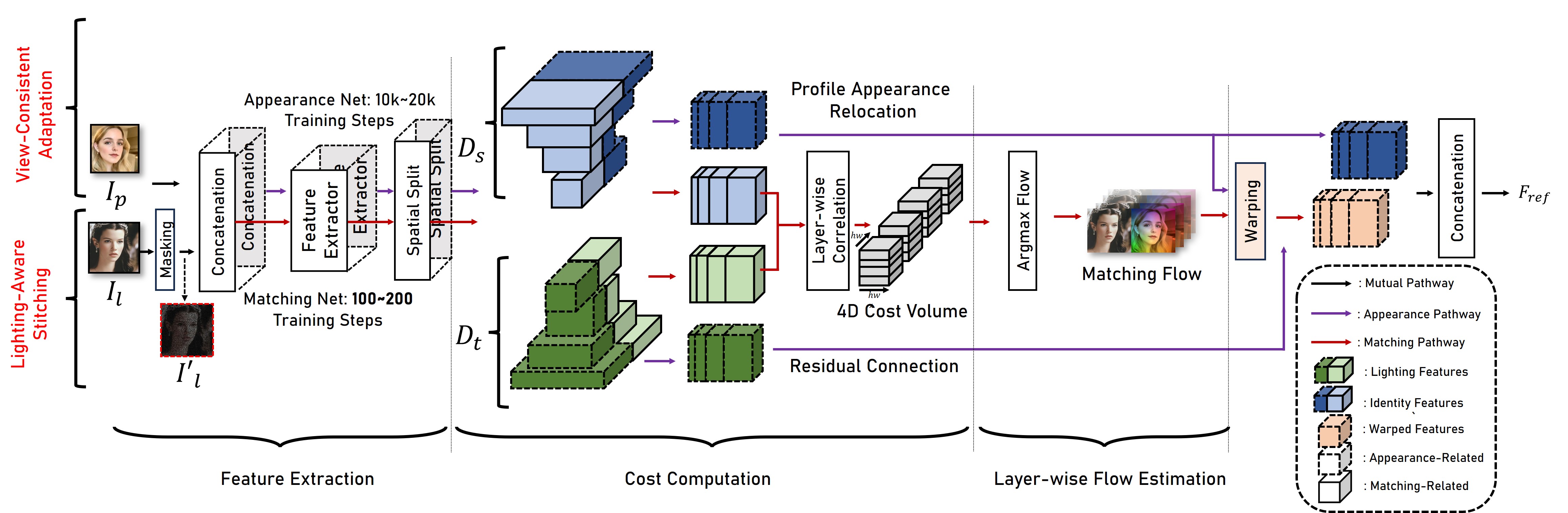}
   \caption{Two main modules as feature extractors are presented here: matching net (gray box) and appearance net (white box). The spatially concatenated input $l_p$ and $I'_l$ are fed into these separately trained control UNet to extract multi-level features, where the ones from the matching net are used to compute 4D cost volume for dense matching and warping field, the features from appearance net are used to warp to blend with lighting features. We show the different pathways and feature sources in different color codes.}
   \label{fig:pipeline}
\end{figure*}
\subsection{Data Synthesis Pipeline}
Our data synthesis pipeline consists of three main stages. First, we generate a diverse collection of human face images using the SDXL model, randomly sampling from a curated list of celebrity names as described in \cite{CelebBasis}. Second, we employ an encoder trained to map these generated images into EG3D \cite{EG3D} tri-plane representations, following the reprojection methodology detailed in \cite{Single-Image}. Finally, we leverage these tri-plane representations to render multiple novel viewpoints of each subject. This process yields a comprehensive dataset containing varied perspectives of each individual, which is crucial to develop robust models for different poses and viewing angles.
% Here, we describe the pipeline for data synthesis. Our approach begins with generating a diverse set of human images using SDXL by randomly selecting celebrity names, following the procedures outlined in \cite{CelebBasis}. These initial images serve as the foundation of our data synthesis. Subsequently, we reproject the generated images into the latent space using a technique detailed in \cite{Single-Image}. This involves training an encoder that maps the images to EG3D \cite{EG3D} tri-planes and then re-renders them to obtain novel views. By doing so, we create a comprehensive dataset that includes multiple views of the same subject, enriching the data with different perspectives, which is essential for training models that must handle a variety of poses and orientations.

% For each pair of generated multi-view images, we randomly select one as the image reference. Then, we randomly drop pixels in the other image, as depicted in Fig. \ref{fig:pipeline} (left). This random pixel-dropping step is crucial for constructing control conditions, simulating different levels of light preservation, and training our model to handle the identity-lighting trade-off effectively.

\subsection{In-Context View $\leftrightarrow$ Lighting Consistency}
% \subsection{Simple Implementation}
IC-Portrait aims to generate ControlNet-like~\cite{controlnet} reference features highly compatible with existing ControlNet-family models, facilitating easy integration into existing pipelines. This design choice is motivated by the widespread adoption and effectiveness of ControlNet-based architectures in various image generation tasks~\cite{DiffPortrait3d,ControlNext,ControlVideo}. We utilize the same synthetic multi-view dataset for both the appearance and matching net training, as in Fig.~\ref{fig:pipeline} with the only difference that the matching net is trained with only $100\sim200$ steps. This unified data source simplifies the training process, as we do not need to perform excessive hyperparameter tuning for different settings. 

Although we conceptually decompose portrait generation into two tasks, namely Lighting-aware Stitching and View-Consistent Adaptation, the View $\leftrightarrow$ Lighting consistency is learned in an end-to-end simultaneous manner. This means that rather than training the model on these tasks one by one, we incorporate them into an in-context framework where the two conditional inputs are concatenated horizontally. The noised latents are connected at a spatial level (along the width dimension) to be fed into the diffusion UNet. 

The training objective of our IC-Portrait framework is designed to be simple yet effective. Let $z$ and $z'$ be a sampled view pair from our multi-view dataset, and let $c_f$ be the input condition, which is obtained through our high-proportion masking and concatenation process. Given a time step, we use $z'$ as the CLIP \cite{CLIP} image embedding prompt (without using a text prompt). The image diffusion algorithms learn a network (UNet) to predict the noise added to it.

\noindent \textbf{Masking.} 
The masking technique employed in our framework is inspired by Masked Autoencoders (MAE) \cite{MAE}. Instead of using traditional sampling methods like checkerboard sampling, which often leads to overly smooth results due to its uniform sampling grid, we adopt a more sophisticated sampling strategy. Our sampling method introduces complex spatial variations in the image, avoiding trivial convergence during training. Specifically, we randomly sample pixels to enable the model to learn from masked and unmasked regions, thereby enhancing its ability to handle identity and lighting information simultaneously.

\begin{figure}[thb]
  \centering
   \includegraphics[width=1\linewidth]{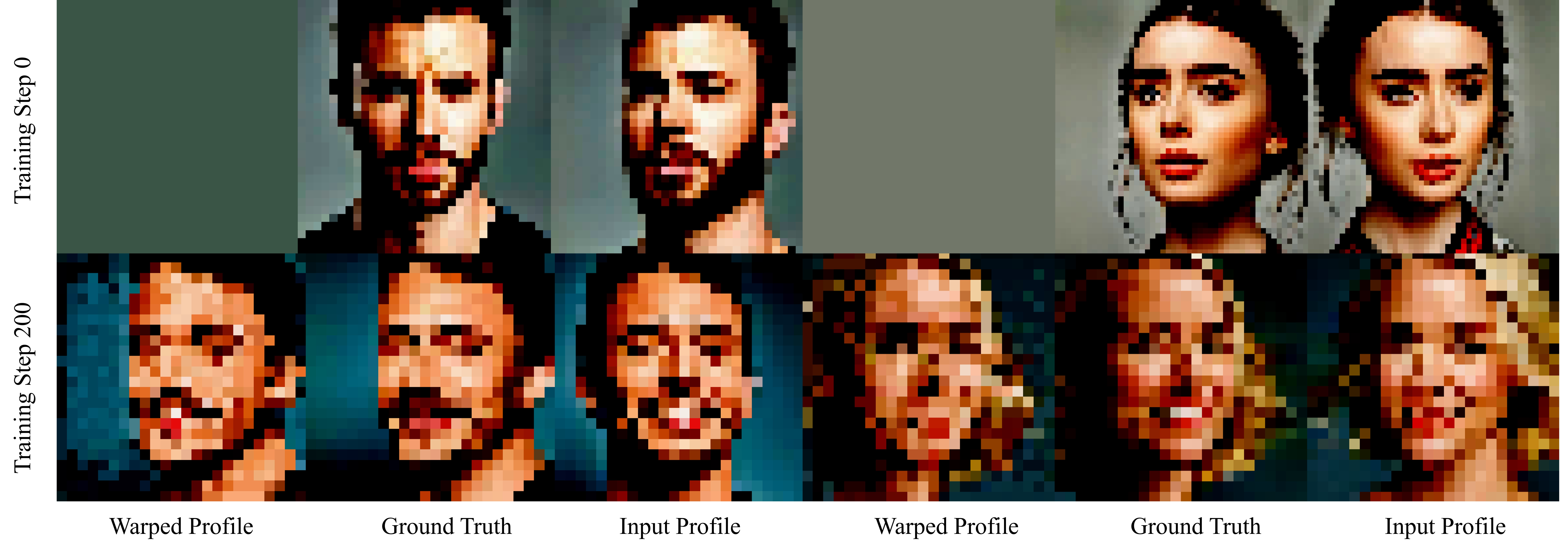}
   \caption{When initializing the matching net with a pre-trained UNet (0 step), it fails to generate meaningful matching features. However, after just 200 training steps, the matching net acquires the ability to promise a reasonable matching field. }
   \label{fig:correspondence}
\end{figure}

 Consider a sampled view pair $z$ and $z'$. We apply a sampling function $F(\cdot,n)$ that uniformly samples $n\%$ of the pixels in the image. Given a sampled view pair $z$ and $z'$, we obtain the input condition $c_f$ as:
\begin{equation}
    c_f = F(z, 0\sim n) \textcircled{c} z',
\end{equation}
where $\textcircled{c}$ denotes spatial width-wise concatenation, and $0\sim n$ represents a value uniformly picked within the range $(0,n)$. The input data $z_0$ is then defined as:
\begin{equation}
    z_0 = z\textcircled{c}z'.
\end{equation}

Then the training objective $\mathcal{L}$ is formulated as follows:
\begin{equation}
\mathcal{L}=\mathbb{E}{z_0,t,z',c_f,\epsilon\sim\mathcal{N}(0,1)}\left[ \lVert\epsilon-\epsilon_\theta(z_t,t,z') \rVert_2 \right].
\end{equation}

\noindent \textbf{In-Context Matching.} In order to fully preserve the profile facial details, we try to use an explicit way to model the correspondence. Pre-trained diffusion UNet~\cite{LatentDiffusion} already inherently preserve the in-context self-attention reference capability, we want to further improve this ability explicitly, by adopting ideas from dense matching~\cite{LearningOpticalFlow}. 
Generally, to find the matching field $F^{*}$ that maximizes the posterior, we can use the maximum a posteriori (MAP) approach, following the formulation in \cite{DiffMatch}:
{\footnotesize
\begin{equation}
\begin{aligned}
F^{*} &= \underset{F}{\mathrm{argmax}}\, p(F|D_{\mathrm{src}},D_{\mathrm{tgt}}) = \underset{F}{\mathrm{argmax}}\, p(D_{\mathrm{src}},D_{\mathrm{tgt}}|F) \cdot p(F) \\
&= \underset{F}{\mathrm{argmax}}\{
\underbrace{\log p(D_{\mathrm{src}},D_{\mathrm{tgt}}|F)}_{\textrm{data term}}
+\underbrace{\log p(F)}_{\textrm{prior term}}
\},
\end{aligned}
\end{equation} }

where $D_{\mathrm{src}}$ and $D_{\mathrm{tgt}}$ are the feature descriptors. While classical dense matching pipelines~\cite{cats,PDC-Net+,GLU-Net,GOCor}, use off-the-shelf feature extractors, we use the diffusion feature space to extract multi-level semantics as feature descriptors. However, the diffusion features cannot be directly adopted to new input conditions, \textit{i.e.}, masked image concatenated with profile reference. In practice, $100\sim 200$ training steps suffice to tune the matching net to produce meaningful descriptors. We hypothesize that the self-attention mechanism inside the matching net, initialized with diffusion UNet, works as a matching prior term, making such fast convergence possible.

\noindent \textbf{Flow Computation.} We utilize a comprehensive flow computation following~\cite{Dreammatcher} to compute the flow between different views and lighting conditions. Let $I_\mathrm{src}$ and $I_\mathrm{tgt}$ be two images from our dataset representing different views or lighting conditions. The cost volume $C(u,v)$ between them can be computed using a cost volume-based method, which is built upon the following equation:
\begin{equation}
\mathcal{C}(u,v)=\frac{D_{\mathrm{src}}(u)^TD_{\mathrm{tgt}}(v)}{ \left\lVert D_{\mathrm{src}}(u)\right\rVert_2 \left\lVert D_{\mathrm{tgt}}(v) \right\rVert_2},
\end{equation}

Here, $D_{\mathrm{src}}(u)$ and $D_{\mathrm{tgt}}(v)$ denote feature vectors extracted from the source image and target image at spatial locations $u$ and $v$, respectively, at time step $t$. These feature vectors are obtained from an intermediate layer of the matching UNet, as shown in Fig.~\ref{fig:pipeline}.

The cost volume $\mathcal{C}$ quantifies the similarity between feature vectors at different locations in the two images. To obtain the matching field $F$, we perform an \textit{argmax} operation over $\mathcal{C}$ to find the best matching locations. Mathematically, for each pixel in $I_\mathrm{src}$, we determine its most likely corresponding location in $I_\mathrm{tgt}$ by:
\begin{equation}
F=(dx(u, v), dy(u, v)) = \text{argmax}_{(dx, dy)} \mathcal{C}(u + dx, v + dy),
\end{equation}
where $F$ is the matching flow and $dx$, $dy$ are the offset field. We visualize the correspondence in Fig.~\ref{fig:correspondence}.

\noindent \textbf{Annealed Feature Aggregation.} The multi-level features derived from the identity profile $I_p$ are warped using the estimated matching flow $F$. To directly incorporate this information, we blend the warped feature with the lighting feature. We then recompose these two features by concatenating them along the spatial width dimension.

For the original lighting features, we combine them with the warped features in a residual connection setup. Since we are performing these operations at multiple levels, we implement a non-uniform weighting strategy for merging the weights.

Our aim is to make the model pay more attention to spatial details. Thus, we use a linear annealing schedule to blend the features. We assign larger weights to the low-level features that are closer to the input. Mathematically, let $W_{l}$ be the weight for the $l$-th level of features, where $l$ represents the level number in the multi-level structure. The weights are determined as follows:
\begin{equation}
 W_{l}=(1-\frac{l}{L})\times\alpha+\beta,  
\end{equation}

where $L$ is the total number of layers in the UNet~\cite{LatentDiffusion} down-sample blocks, $\alpha$ and $\beta$ are constants that control the rate of annealing and the base weight, respectively. This formula ensures that as $l$ (the level number) increases, the weight $W_{l}$ decreases linearly, giving more importance to the lower-level features. This way, the model can better capture the fine-grained spatial details present in the lower-level features while still integrating information from higher-level features.

\noindent \textbf{Progressive Inference}. The full inference of IC-Portrait employs a progressive generation scheme. The rationale behind this lies in the inherent properties of the diffusion process, which involves corrupting by noise and subsequent rebuilding. In this process, we conduct input projection by adding Gaussian noise to human faces. Our aim is to match the distribution of another identity under identical lighting and shading conditions. By carefully selecting and adding an appropriate amount of noise, we can maximize the retention of the original lighting while simultaneously effecting the desired identity transformation. However, it is crucial to note that adding either excessive or insufficient noise is not optimal.

% In terms of implementation, we progressively introduce an appropriate quantity of noise to the human faces in a controlled fashion, leveraging the natural characteristics of the diffusion process. Once the noise has been added, we then proceed with the denoising of the noisy image. Through this sequential process of noise addition and denoising, we are able to generate more balanced results. This progressive approach strikes a balance between preserving the original lighting and effecting the necessary identity change. It enables us to gradually shift the style reference face towards the desired identity while ensuring that the lighting conditions remain consistent throughout the transformation. This gradual transition prevents sudden alterations that might otherwise lead to the loss of critical features, thereby avoiding an unbalanced final output.

% Some commands used in this file
\newcommand{\package}{\emph}

\begin{figure*}[thb]
  \centering
   \includegraphics[width=0.95\linewidth]{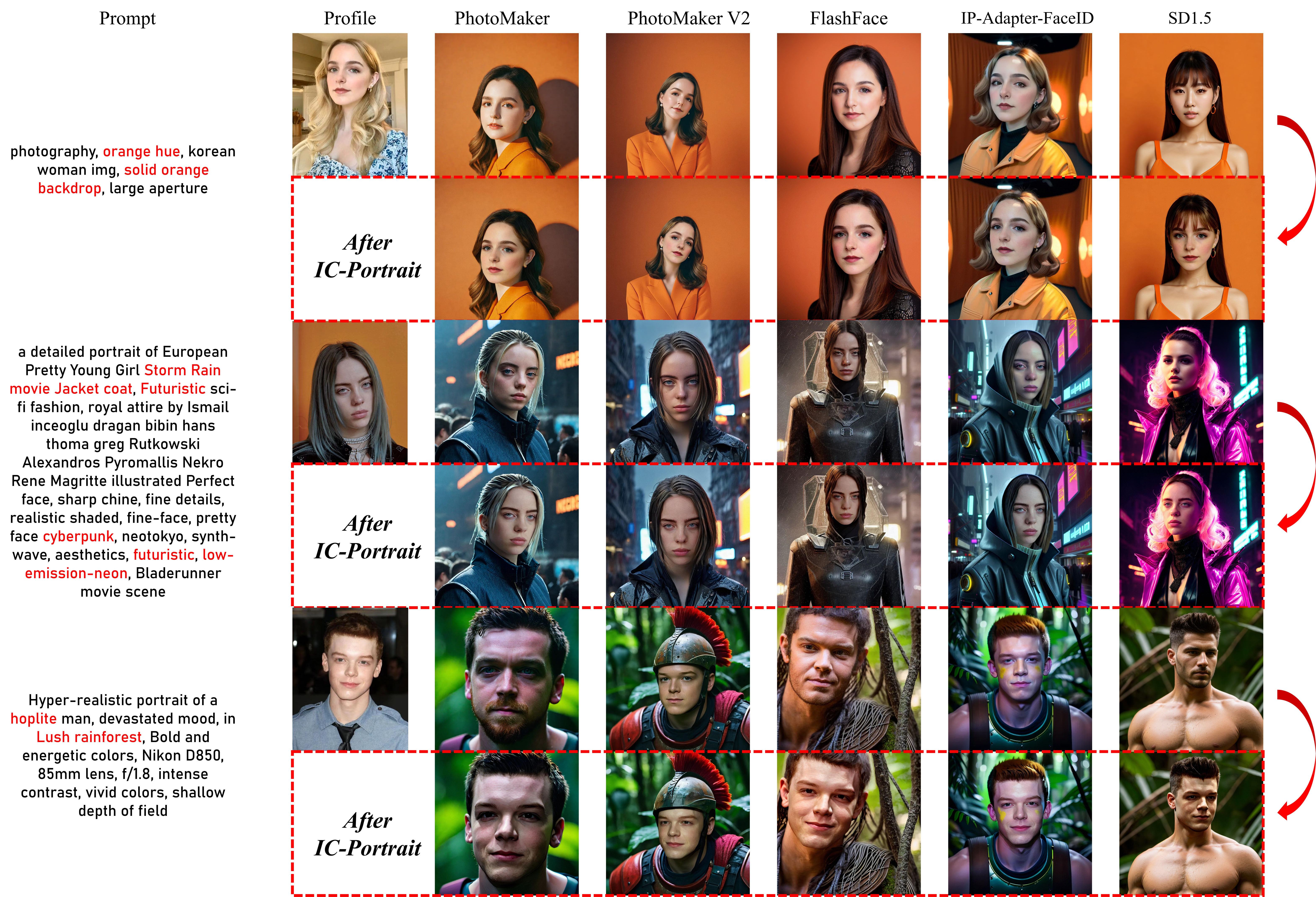}
   \caption{We pick the current state-of-the-art portrait generation methods as baselines. Since our setting needs a style reference, we mimic the text-based setting by first generating a style reference with SD1.5 and running IC-Portrait afterward. We also present a new comparison setting where we use the generated results from baseline methods as style reference and refine the identity with IC-Portrait to clearly show the superiority of our method in terms of ID similarity and coherence.}
   \label{fig:visual_comparison}
\end{figure*}
\section{Experiments}
\begin{figure}[htb]
  \centering
   \includegraphics[width=0.7\linewidth]{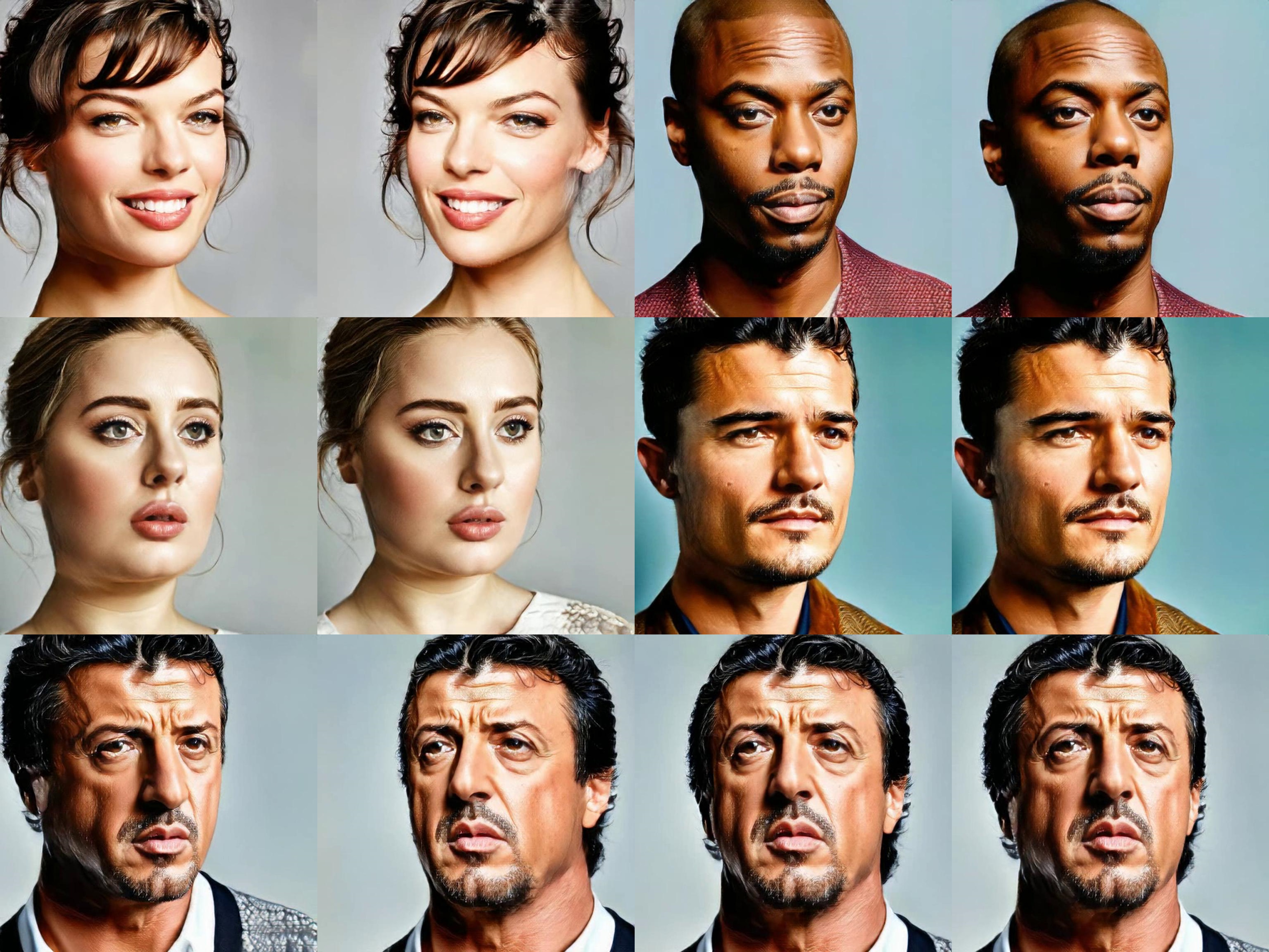}
   \caption{The synthetic multi-view dataset. We employ SDXL to generate human images by randomly selecting celebrity names. Then, we reproject the image into the latent space and generate multi-view images using EG3D.}
   \label{fig:multi-view}
\end{figure}

% \section{Setup}
\noindent \textbf{Datasets.}~Following the dataset synthesis process in CelebBasis~\cite{CelebBasis}, we use celebrity names to trigger the identity generation to curate our dataset. We use 102 male celebrities such as ``Leonardo DiCaprio", ``Brad Pitt", etc., and 104 female celebrities such as ``Scarlett Johansson", ``Jennifer Lawrence", etc.
% The template of prompts is given as: 
% {\it  ``a handsome man \{Random.choice (celebrity\_names)\} front-view portrait with \{Random.choice (expression)\}, detailed face, different clothes, raw photo"}. We give two expressions: ``smile" and ``neutral face" since we find these two expressions show the clearest identity.
After generating a front-view image with the given prompt using SDXL~\cite{SDXL}, we use one-shot EG3D-based novel-view synthesis technique~\cite{Single-Image} which inverts the portrait image into tri-plane NeRF~\cite{NeRF} representations so we can do re-renderings by changing the 25 scalars of camera parameters as in \cite{EG3D}. We randomly sample two views from the novel-views to construct pair data, which finally make up a training set of 35200 pairs as shown in Fig.~\ref{fig:multi-view}. 
% Two important factors to affect the quality of this synthetic dataset are the ID diversity and perspective diversity. 
% To have a diverse gallery for learning general appearance, we need to have enough portrait IDs, which requires us to give enough celebrity names to trigger the generation. The perspective diversity is guaranteed by the volume rendering nature of NeRF based method so that we can sample as dense as we want.

\noindent \textbf{Implementation Details.}~Our model is based on StableDiffusion-v1.5~\cite{LatentDiffusion} model for the matching and appearance net. The reference features in ControlNet~\cite{controlnet} format are compatible with any ControlNet-friendly framework. To allow a more straightforward formulation, we remove the text prompt of UNet and ControlNet with compatible CLIP~\cite{CLIP} ViT-L/14 embedding. 1)
In the matching and appearance training phase, we initialize our model by copying the weights of pre-trained UNet following \cite{controlnet}. We fix the UNet and clip encoder during training and only train the reference network. We first train the matching net and fix it to provide matching guidance for appearance net training. We train our appearance net for 20k steps on 4 NVIDIA A40 GPUs, with a constant learning rate of $1e^{-5}$ and a batch size of $16$, while 200 steps for matching net. We select one of the pixel sampling images as either right or left and use the other as the profile reference. We set the embedding as a zero tensor with a 10\% chance to enable classifier-free guidance. 2) For the DreamBooth~\cite{Dreambooth} customization phase, we align the faces of the profile images first. Then, we train the whole UNet for customization to maximize identity preservation at learning rate 2e-6 and batch size 1 on a single NVIDIA A40 GPU.

\noindent\textbf{Evaluation Metrics.}~We assess the image generation quality mainly regarding identity preservation since IC-Portrait supports reference-image-based portrait generation. We do not need to care about text-image alignment in our generation. We use arcface~\cite{arcface} as ID embedding extractor and cosine similarity as our similarity metric. We use 11 identities (all celebrities) images as reference images with 40 unique prompts to generate for each profile image. For the Arcface similarity, the higher is better, and the upper bound is 1. We compute the similarity between all the profile images for each result and compute the statistics as in Tab.~\ref{tab:quant}.

\subsection{Qualitative Comparisons}
\label{sec:qualitative}

To evaluate the performance of IC-Portrait, we conduct extensive qualitative experiments. We compare our method with several state-of-the-art portrait generation techniques: PhotoMaker~\cite{photomaker}, PhotoMaker v2~\cite{photomaker}, FlashFace~\cite{flashface} and IP-Adapter-FaceID-plus (IPA-FaceID)~\cite{IP-Adapter}. As shown in Fig. \ref{fig:visual_comparison}, to enable a comparison with the text-driven generation, we first generate style references using StableDiffusion-v1.5 (SD1.5) for a fair comparison with traditional text-based methods; then, we apply IC-Portrait to the generated style reference image.

The results clearly demonstrate that our method outperforms the baselines regarding ID similarity while achieving comparable coherence and photorealism. In addition, we introduce a unique secondary comparison approach. We take the outputs of the baseline methods as style references and use IC-Portrait to refine the identity. This novel comparison showcases the remarkable ability of IC-Portrait to enhance identity preservation, producing portraits that are visually more appealing and true to the subject's identity.

\begin{figure*}[htb]
  \centering
   \includegraphics[width=0.7\linewidth]{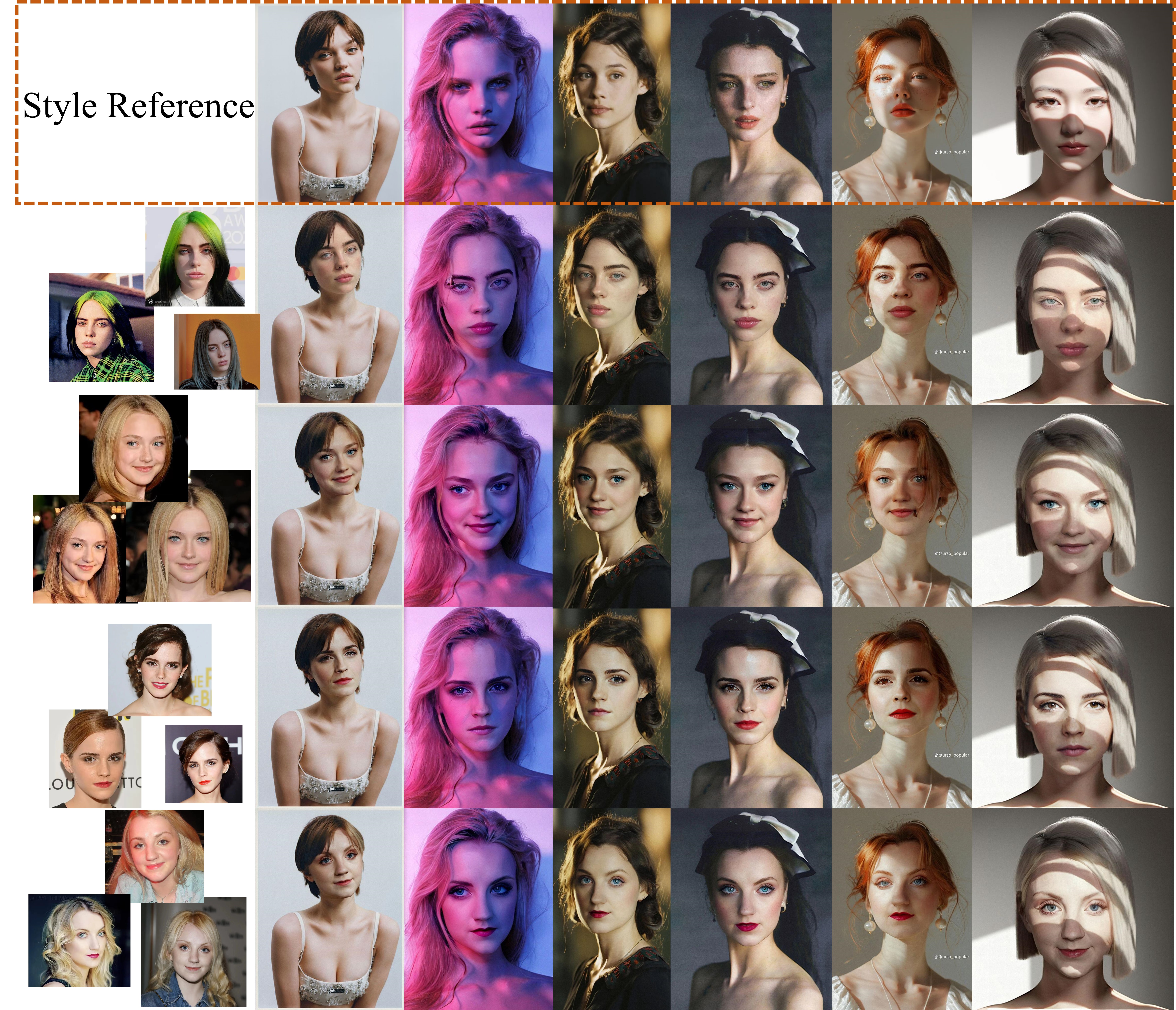}
   \caption{We show the results of IC-Portrait under extreme lighting conditions.}
   \label{fig:stronglighting}
\end{figure*}

From the results in Fig.~\ref{fig:visual_comparison}, PhotoMaker series give high-quality results with good prompt-alignment but fail to preserve the fine details of the profile facial features. IP-Adapter-FaceID and FlashFace demonstrate the best ID preserving abilities and image quality but still only depict a rough resemblance just as I said in Fig.~\ref{fig:diverse}; each only depicts some profile characteristics rather than fully preserves them. In the meantime, IC-Portrait can completely maintain the identity at a novel-view level but still achieves coherence and photorealism. The results after applying IC-Portrait can show how the face is changed, making it clear that “This is the same person”. 
% It's as if we first reposition the face, relight it according to the style reference, and then paste it back to the face region, but in a more end-to-end and inherent manner.
We also show in Fig.~\ref{fig:stronglighting} that IC-Portrait can handle extreme lighting conditions.
% From the extremely diverse and out-of-distrubution lighting conditions of the style reference images IC-Portrait demonstrates consistently good performance credited to our dual condition ID-ReferenceNet.

\subsection{Novel-View Synthesis and Relighting Capabilities}

\begin{figure}[thb]
  \centering

   \includegraphics[width=1\linewidth]{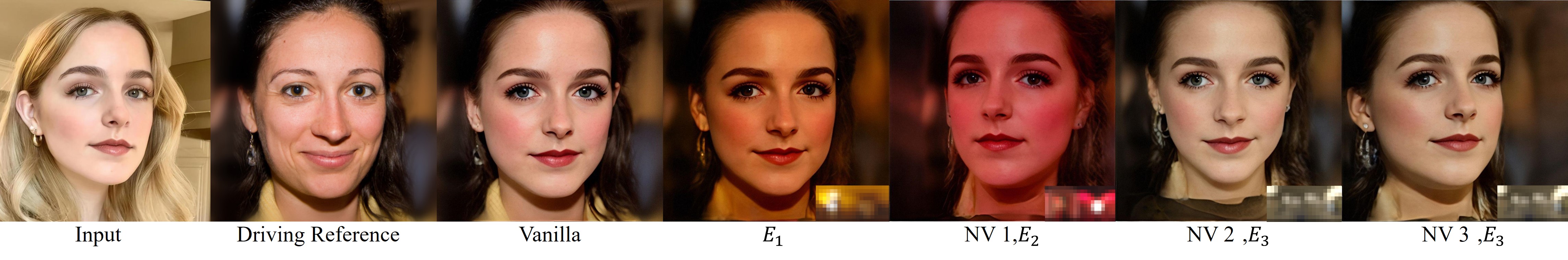}
   \caption{IC-Portrait effectively transfers and relights identities across various conditions, preserving facial characteristics and subject integrity while adapting to new poses and lighting from a driving reference.
   % We use the column 1 input as a profile and transfer it onto the driving reference image to get the vanilla setting result in column 3. Novel Views (NV) are generated with different environment maps ($E$). We adopt an off-the-shelf novel-view synthesis and relighting method to reposition and relight the driving reference image and apply IC-Portrait to it. We can see the ID consistency under different views and environment map lighting conditions with a certain profile photo.
   }
   \label{fig:novel_relight}
\end{figure}

As depicted in Fig. \ref{fig:novel_relight}, we transfer an input profile from column 1 onto the driving reference image. The resulting vanilla setting in column 3 provides a baseline for comparison. When generating novel views (NV) with different environment maps ($E$), we use an off-the-shelf novel-view synthesis and relighting method to generate relighted and reposed portraits. We then re-render it to the target identity and thus produce such effect as demonstrated. This allows us to observe the consistent preservation of identity under diverse viewing angles and various lighting conditions imposed by the environment maps. It showcases the adaptability of IC-Portrait in handling complex visual scenarios such as lighting and perspective changes while maintaining the integrity of the person's identity.

% In the previous sections, we showed IC-Portrait's inherent ability to conduct portrait generation and apply novel-view synthesis and relightable stitching. Here, we want to directly demonstrate the explicit power and flexibility of our method in novel-view synthesis and relighting. To achieve this, we conducted a series of experiments and comparisons to clearly showcase IC-Portrait's unique capabilities. 
% NeRF-based methods such as \cite{Single-Image, GOAE,Triplanenet} use newly trained encoders to map the input image to its corresponding tri-plane NeRF representation to render novel views via volumetric rendering. Diffusion-based methods such as \cite{DiffPortrait3d} use attention injection to strongly condition the generation process according to the reference view. \cite{lite2relight} can perform 3D-aware relighting while generating novel-views by training an EG3D-based encoder on light stage data, but light stage data can be extremely expensive.
% IC-Portrait can achieve similar results with synthetic-only data while achieving 3D novel-view synthesis. Although our setting is not strictly the same as the aforementioned methods, we here demonstrate the possibilities and advantages of our approach. By leveraging our carefully designed pipeline and techniques, IC-Portrait is able to generate high-quality novel views and perform effective relighting with a more accessible and cost-effective approach. This not only shows the practicality of our method but also paves the way for more efficient and flexible portrait generation in various applications.

\subsection{Quantitative Experiments}

We mainly focus on ID similarity metrics since IC-Portrait aims at reference-based portrait generation rather than text-based. To fairly compare our method with other state-of-the-art, we use Stable Diffusion-v1.5 (SD1.5) \cite{LatentDiffusion} to generate reference images with our test prompts just as the baseline methods do and run IC-Portrait afterward. The metrics are given in Tab.~\ref{tab:quant}. IC-Portrait achieves the highest cosine similarity compared to baseline methods in all statistics. 

\begin{table}[htbp]
\centering
\caption{Metrics Comparison. We use cosine similarity of arcface~\cite{arcface} embeddings. The ground-truth (Upper Bound) is derived from the pair-wise similarity of different profiles of the same person in our experiment.}
\begin{tabular}
{c|c|c|c|c}
\hline
Method & Min & Max & Median & Mean \\
\hline
Upper Bound & 0.504 & 0.828 & 0.655 & 0.658 \\
\hline
\footnotesize PhotoMaker~\cite{photomaker} & 0.271 & 0.423 & 0.354 & 0.351 \\
\footnotesize  PhotoMaker v2~\cite{photomaker} & 0.360 & 0.574 & 0.444 & 0.453 \\
 \footnotesize CustomDiffusion\cite{custom_diffusion} & 0.347 & 0.500 & 0.432 & 0.429 \\
\footnotesize  DreamBooth~\cite{Dreambooth} & 0.398 & 0.571 & 0.487 & 0.487 \\
\footnotesize  FlashFace~\cite{flashface} & 0.383 & 0.649 & 0.464 & 0.479 \\
\footnotesize  IPA-FaceID~\cite{IP-Adapter} & 0.377 & 0.649 & 0.469 & 0.480 \\
\hline
Ours w/o annealing & 0.466 & 0.663 & 0.550 & 0.553 \\
Ours & \textbf{0.479} & \textbf{0.674} & \textbf{0.562} & \textbf{0.565} \\
\hline
\end{tabular}
\label{tab:quant}
\end{table}

IC-Portrait can achieve such excellent results because of its unique design and advanced techniques. The novel-view-synthesis nature helps to copy the identity and appearance of the profile image completely and thus precisely preserve the unique characteristics of the subject. By leveraging advanced 3D-aware novel-view synthesis techniques and understanding facial geometries, it can re-adjust the user's profile photo to fit the pose of the style reference image in a highly accurate manner.

The baseline methods rely on rough representations such as ID embedding, and these fall short of capturing the fine details and nuances of the subject's identity. They struggle to handle the complex variations in user images caused by makeup, lighting, and expressions, resulting in less accurate and less similar portraits than IC-Portrait. Additionally, the iterative dual condition inference process in IC-Portrait allows for a more refined transfer of the reference ID to the target image while maintaining the integrity of the lighting and other visual details, further enhancing the overall quality and ID similarity of the generated portraits.

Besides, the \textit{Min} result of ground-truth (Upper Bound) (Tab.~\ref{tab:quant} row~1) is only around 0.5, which shows the great gap within the profiles of the same person. This statistic strengthens the main motivation of IC-Portrait-the profile diversity problem.

\subsection{Handling Multiple Subjects
}
\begin{figure*}[htb]
  \centering

   \includegraphics[width=0.7\linewidth]{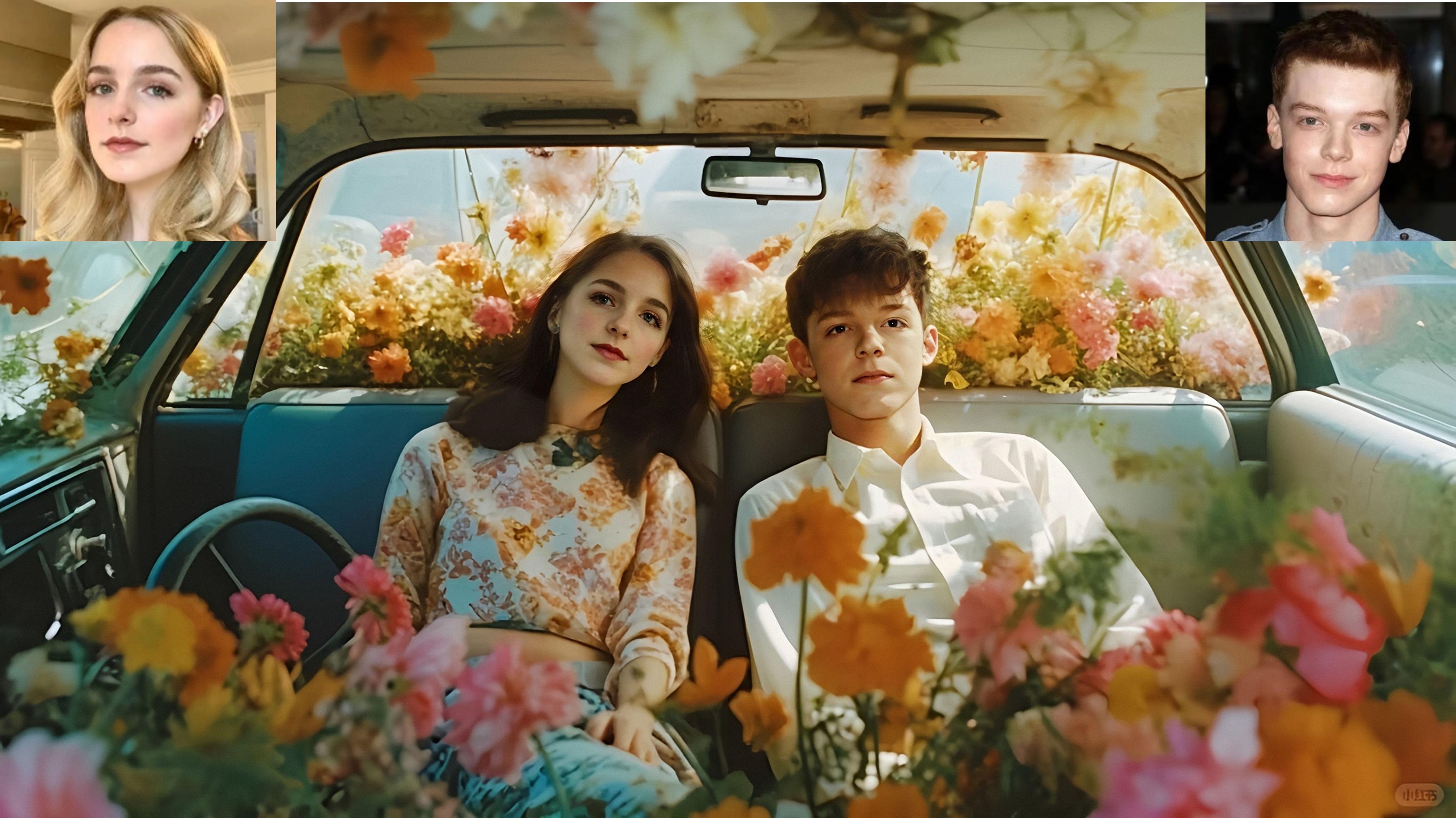}
   \caption{Here we show the multi-person setting of IC-Portrait. IC-Portrait inherently supports multi-person setting by processing faces sequentially.}
   \label{fig:doubleshot}
\end{figure*}

In the context of multiple people settings, IC-Portrait demonstrates its versatility. As seen in Fig. \ref{fig:doubleshot}, our method processes multiple faces in a sequential manner. This sequential processing mechanism ensures that each individual's identity is accurately captured and integrated within the group portrait. Since the processing of each face is completely separate, we solve the ID mixing problem completely. FastComposer~\cite{FastComposer} and UniPortrait~\cite{Uniportrait} try to use attention localization/routing method to determine the region of attention influence but still have difficulty in completely maintaining individual identity. This capability makes IC-Portrait highly suitable for applications involving group portraits or scenes with multiple subjects.

% \section{Self-Swapping and Identity Preservation}

% \begin{figure}[htb]
%   \centering
%    \includegraphics[width=1\linewidth]{images/self-swap.jpg}
%    \caption{Here we demonstrate a self-swapping experiment within a user's profiles to show the novel-view nature of our method. Since the profiles are diverse in expressions and makeup, with IC-Portrait we entirely preserve the characteristics of that specific profile photo while reposing it, just like novel-view synthesis.}
%    \label{fig:selfswap}
% \end{figure}
% The self-swapping experiment, as illustrated in Fig. \ref{fig:selfswap}, highlights an interesting aspect of IC-Portrait. Given the diversity in expressions and makeup within a user's profiles, our method can perform self-swapping while preserving the specific characteristics of each profile photo. This process is analogous to novel-view synthesis, where the pose and appearance of the subject are adjusted while maintaining the core identity. It showcases the flexibility of IC-Portrait in handling variations within a single user's data, further emphasizing its ability to generate personalized and accurate portraits.

\subsection{Ablation Study}

\begin{figure}[thb]
  \centering
   \includegraphics[width=1\linewidth]{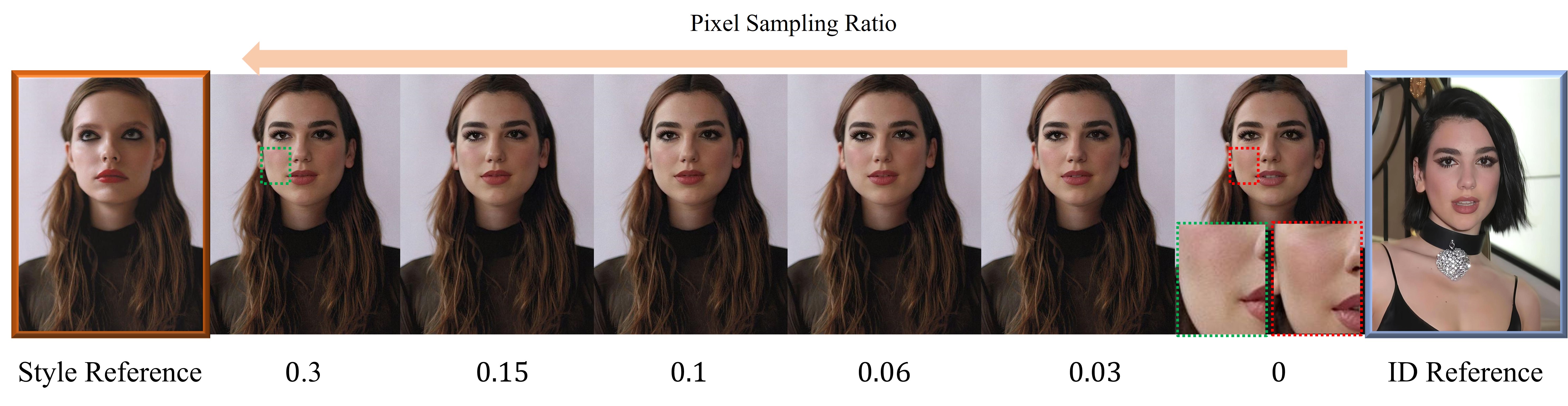}
   \caption{Higher sampling ratios result in improved shading alignment with the style reference (left), as seen in the highlighted areas of the zoom-in results, but may introduce a trade-off with generation stability.}
   \label{fig:pixel_sample}
\end{figure}

We show the pixel sampling ratio ablation in Fig. \ref{fig:pixel_sample}. As the core motivation in IC-Portrait, we formulate our idea from MAE \cite{MAE}. The more dense the sampling, the more clearly we can observe that the shading aligns better with the original style reference. By presenting the zoom-in results of the face region for comparison, it becomes evident that with a denser sampling ratio, elements such as skin color, especially in the highlighted areas, match the style reference more closely.
% Interestingly, the identity remains relatively stable even as the sampling ratio increases. However, it is important to note that there is a trade-off, as the stability of the generation process may degrade during actual usage.

We also show that without our annealed feature aggregation, the identity-preserving is not as good as our full model as in Tab.~\ref{tab:quant}.

% This formulation actually provides strong insights into the design of our dual-condition approach. We can manipulate the pixel sampling ratio to balance between maintaining identity and achieving better alignment with the style reference. By carefully adjusting this parameter, we can optimize the visual quality of the generated portraits. IC-Portrait is thus able to offer a more refined control over the generated output, enabling users to obtain portraits that not only preserve their unique identity but also blend seamlessly with the desired style reference. This fine-grained control over pixel sampling further solidifies IC-Portrait's position as an advanced and versatile tool in the portrait generation domain, allowing for more precise customization and better adaptation to various user preferences and application scenarios.

\section{Conclusion}

IC-Portrait presents a novel approach to personalized portrait generation, adeptly handling the challenges of diverse user-profiles and varied lighting conditions by decomposing the task into \textbf{lighting-aware stitching} and \textbf{view-consistent adaptation}. By leveraging a masked autoencoding technique and a synthetic multi-view dataset, the framework learns in-context correspondence that allows it to accurately transfer both identity and stylistic elements from reference images, resulting in high-fidelity portraits that are both view-consistent and responsive to diverse lighting scenarios. Extensive evaluations demonstrate that IC-Portrait outperforms existing state-of-the-art methods in terms of identity preservation and visual quality, highlighting its potential for various applications in digital content creation, virtual avatars, and beyond, while also paving the way for future research into more explicit semantic matching and enhanced novel-view synthesis.

{\small
\bibliographystyle{ieee_fullname}
\bibliography{sample-bibliography}
}

\end{document}